\documentclass{arxiv}
\usepackage[american]{babel}

\usepackage{times}

\usepackage[utf8]{inputenc} 
\usepackage[T1]{fontenc}    
\usepackage{hyperref}       
\usepackage{url}            
\usepackage{booktabs}       
\usepackage{amsfonts}       
\usepackage{nicefrac}       
\usepackage{microtype}      
\usepackage{xcolor}         

\usepackage{amsmath}
\usepackage{amssymb}
\usepackage{physics}
\usepackage{mathtools}
\usepackage{xfrac}
\usepackage{setspace}
\usepackage{pgf}
\usepackage[ruled]{algorithm2e}
\usepackage[labelsep=period]{caption}
\usepackage{wrapfig}
\usepackage{placeins}

\newtheorem{assumption}{Assumption}

\DeclareMathOperator{\E}{\mathbb{E}}

\DeclareMathOperator{\diff}{\mathrm{d}\!}

\ifdefined\nohyperref\else\ifdefined\hypersetup
  \definecolor{mydarkblue}{rgb}{0,0.08,0.45}
  \hypersetup{ %
    pdftitle={},
    pdfsubject={},
    pdfkeywords={},
    pdfborder=0 0 0,
    pdfpagemode=UseNone,
    colorlinks=true,
    linkcolor=mydarkblue,
    citecolor=mydarkblue,
    filecolor=mydarkblue,
    urlcolor=mydarkblue,
    }
  \fi
\fi

\title[A Preprint]{Ensembled Prediction Intervals for Causal Outcomes\\Under Hidden Confounding}
\theseauthors{
  \Name{Myrl G. Marmarelis} \Email{myrlm@isi.edu}\\
  \Name{Greg {Ver Steeg}} \Email{gregv@isi.edu}\\
  \Name{Aram Galstyan} \Email{galstyan@isi.edu}\\
  \Name{Fred Morstatter} \Email{fredmors@isi.edu}\\
  \addr 
  USC Information Sciences Institute\\
  4676 Admiralty Way\\
  Marina del Rey, CA 90292\\ }

\definecolor{nice-blue}{RGB}{120,120,228}
\definecolor{nice-red}{RGB}{244,102,114}
\definecolor{nice-green}{RGB}{44,165,141}

\begin{document}
\maketitle

\newcommand{\notindep}{\ensuremath{ \mathbin{\not\!\perp\!\!\!\perp} }}
\newcommand{\indep}{\ensuremath{ \mathbin{\perp\!\!\!\perp} }}

\begin{abstract}
  Causal inference of exact individual treatment outcomes in the presence of hidden confounders is rarely possible. Recent work has extended prediction intervals with finite-sample guarantees to partially identifiable causal outcomes, by means of a sensitivity model for hidden confounding.
  In deep learning, predictors can exploit their inductive biases for better generalization out of sample. We argue that the structure inherent to a deep ensemble should inform a tighter partial identification of the causal outcomes that they predict.
  We therefore introduce an approach termed \textbf{Caus-Modens}, for characterizing \textbf{caus}al outcome intervals by \textbf{mod}ulated \textbf{ens}embles.
  We present a simple approach to partial identification using existing causal sensitivity models and show empirically that Caus-Modens gives tighter outcome intervals, as measured by the necessary interval size to achieve sufficient coverage. The last of our three diverse benchmarks is a novel usage of GPT-4 for observational experiments with unknown but probeable ground truth.
\end{abstract}


\section{Introduction}\label{sec:intro}

In order for a regression model to make \emph{causal} predictions, the effect of confounders must be disentangled from the effect of the treatment.
For this reason, causal inference is closely related to the problem of domain shift, since the outcome predictor may be learned on observational data while being expected to perform well on the hypothetical domain with fully randomized treatments. More often than not, the available covariates are imperfect proxies for all the confounders in the causal system. This further compounds the task of causal inference, as the hidden confounders must somehow be taken into account. The best hope in these cases is to produce ``ignorance intervals'' that \emph{partially} identify the causal estimands. The tighter the intervals, the more useful the partial identification, which depends on what can be said about the hidden confounders. 

A sensitivity model~\citep{rosenbaum83} in causal inference is a structural assumption~\citep{manski} about the possible behavior of hidden confounders.
It allows causal estimands to be partially identified as long as the extent of hidden confounding is consistent with the sensitivity model.
The dependence of the treatment assignment on confounders, i.e.\ the \emph{propensity} of treatments, is what makes a study observational rather than a fully randomized experiment.
We consider sensitivity models that bound the \emph{complete} propensity (colloquially, the true propensity of treatment assignments for an individual, taking into account all relevant variables, observed or not) in terms of the \emph{nominal} propensity (based just on observed covariates, allowing it to be estimated by regression.) Sensitivity models of this kind were first introduced by \citet{tan} and have become popular due to their generality and simplicity. The most common setting for these models, in line with Tan's initial formulation, is of binary treatments~\citep{jesson21, kallus19, dorn}, in which the Marginal Sensitivity Model (MSM) bounds the ratio of nominal-propensity odds to complete-propensity odds. When that ratio is unit, and the complete propensity equals the nominal propensity at all points, then the covariates are adequate to explain all the confounders. It is worthwhile to broaden the notion of the MSM in light of recent developments with MSM-like sensitivity models for continuous treatments~\citep{jesson22, marmarelis22} and other nonbinary domains. To accommodate these settings, we consider a general form of sensitivity model. 
In this paper, we explore prediction intervals of causal outcomes due to interventions on the treatment variable, termed \emph{outcome intervals}, that incorporate empirical uncertainties~\citep{jesson20} in addition to the orthogonal concept of hidden-confounding uncertainty. Outcome intervals predict \emph{individual outcomes} of treatments disentangled from confounders, relying on a sensitivity model to guide partial identification in the presence of hidden confounders.

\begin{figure}[!ht]\centering
  \includegraphics[width=0.85\linewidth]{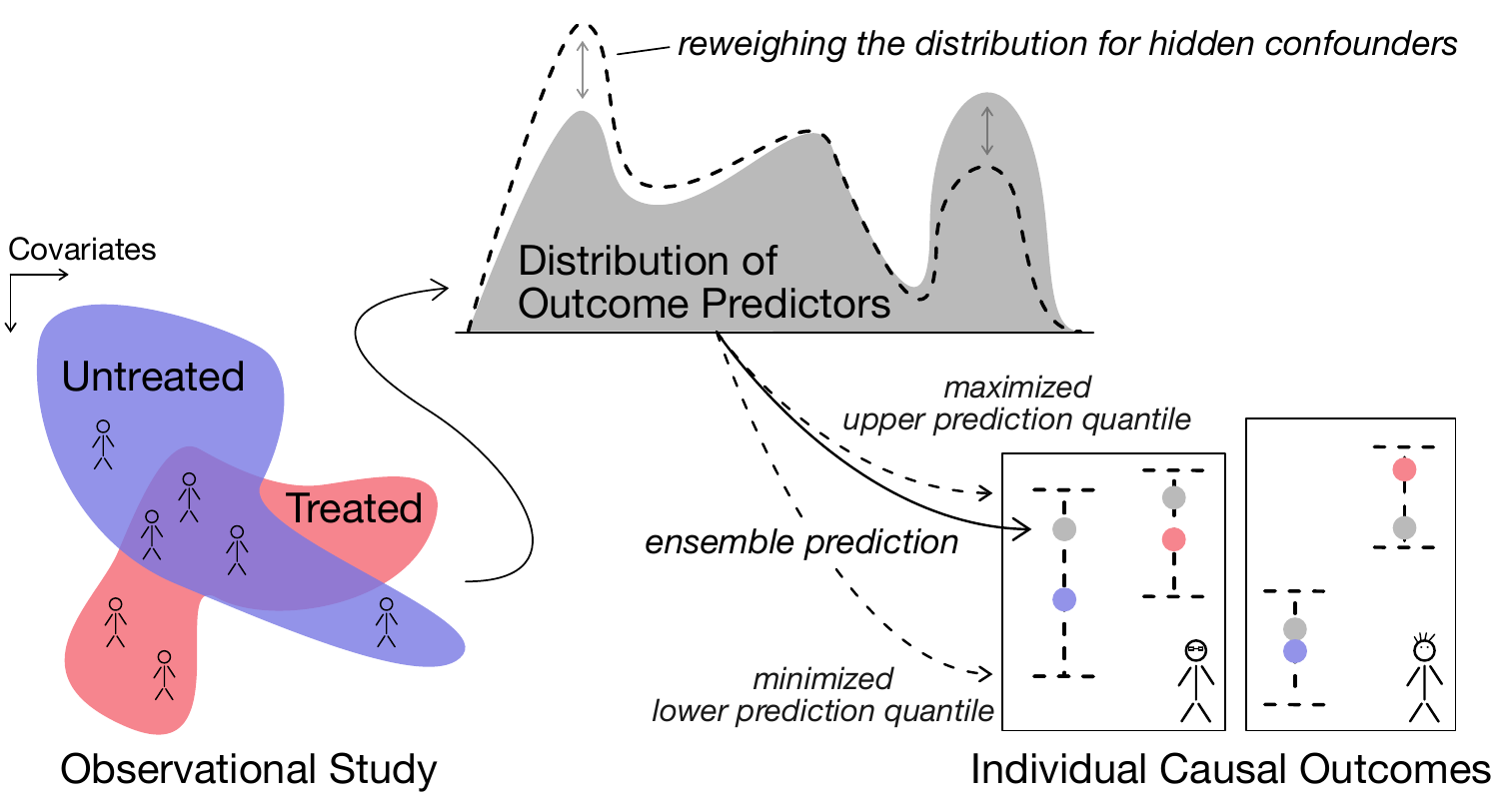}
  \caption{\label{fig:intro}An illustration of the proposed method for causal outcome intervals. First, one samples predictors from a Bayesian posterior or otherwise learns an ensemble to approximate the distribution of outcome predictors that agree with the observational data. The ensemble average (grey dot) could be used to predict actual causal outcomes (red/blue dots). With hidden confounding, the learned ensemble might diverge substantially from the best predictor distribution to model causal outcomes. One cannot identify the correct distribution from observational data alone. Instead, a sensitivity model says how wrong this learned ensemble could be, and one optimizes with respect to weights on the ensemble elements for each individual and treatment in order to upper-bound the $(1-\alpha/2)$ quantile and lower-bound the $(\alpha/2)$ quantile of the ensemble prediction. These intervals incorporate both empirical uncertainties from prediction quantiles and hidden-confounding uncertainties from the ensemble modulation. They are evaluated against ground-truth causal outcomes by removing confounding through interventions on test-set individuals, using (semi-)synthetic data.} 
\end{figure}

\subsection{Related Work}
Our goal diverges from the great strides that have been made in the realm of multiply debiased and robust estimators for \emph{average} outcomes of populations or subpopulations~\citep{athey19, chernozhukov17}. Largely in the binary-treatment context, these estimators have been augmented with \emph{sharp} partial identification methods~\citep{dorn22} that are guaranteed to be valid while not overly conservative. \citet{dorn} accomplishes this partial identification at the cost of having to re-estimate outcome regressions every time the sensitivity model changes.
Partial identification is less explored for nonbinary treatments, which are receiving increased attention~\citep{nie,kaddour,bica20,wang22}. 
Separately, outcome statistics that are more complex than expectations are also of key interest in machine learning~\citep{kallus23}, with diverse purposes like fairness-oriented measures~\citep{kallus22}. Tight partial identification of these other statistics requires novel methodology. Partially identified \emph{outcome quantiles} would be a step in that direction. We solve that problem in this paper for the purpose bounding above and below the individual outcome intervals---our current focus.

The state of the art for partially identified outcome intervals from binary treatments is conformalized~\citep{yin22, jin23}, building on domain shift~\citep{lei21}. Conformal inference looks at the empirical performance of a model to decide how to size its prediction set (interval).
The simplicity of this approach coupled with its finite-sample statistical guarantees makes it widely applicable.
However, conformalized intervals even in the causal setting are based on the behavior of the outcome predictor \emph{on the observed distribution.} This approach is fundamentally limited, especially for causal inference on heavily biased data, which distorts the apparent predictor statistics that conformalization uses to extrapolate to the causal system. We shall discuss these problems and construct an ensemble-based solution with a Bayesian motivation.

\FloatBarrier

\subsection{Motivation}\label{sec:motiv}

Instead of partial identification of (conditional) average treatment effect, (C)ATE, the conformal sensitivity analysis (CSA)~\citep{yin22, jin23} produces rigorous intervals for the individual treatment effect (ITE), in other words the outcome realization rather than expectation.
CSA considers a predictor's performance in a calibration set as a guide for determining prediction intervals out of sample. 
Partial identification tends to be formulated adversarially, in terms of minimizing/maximizing a causal estimand that is admitted by the problem's constraints. CSA involves an optimization problem over the rebalancing weights applied to the calibration sample~\citep{tibshirani,lei21}.
As the conformal method requires quantile estimates, it is impacted by theoretical implications on weighted quantile estimators. The variance of the estimator scales with the variance of the weights~\citep[Theorem~1]{glynn}. If the weights were not inverse-propensity adjusted, then the conformal guarantees would fail due to distribution mismatch, so a large variance from covariate shift, for instance, cannot be avoided.

In place of conformalized domain shift, we posit that an ensemble capturing empirical uncertainties from the observational data could harness its inductive biases to extrapolate to causal outcomes~\citep{jesson20,rame}. 
These elements exist in Bayesian reasoning~\citep{jaynes}, which is a sound and scientific way to reconcile models with data. It incorporates parametric, distributional, structural, and prior knowledge into a \emph{posterior} distribution of learned models that agree with the data. Even with large, deep models that are commonly developed in machine learning, the structural elements of the model contribute to its performance in a general domain~\citep[e.g.][]{edelman22}. In the method presented below, we allow an ensemble's learned biases to aid in extrapolation of the partially identifiable causal estimands. The mathematical connections are clear when the ensemble is supposedly from a Bayesian posterior, but in practice it can be learned in any way that sufficiently captures empirical uncertainties~\citep{wild23}. See Figure~\ref{fig:intro} on the ensemble reweighing.

\section{Approach}\label{sec:appr} 

We present a versatile, modular procedure for taking an ensemble of outcome predictors and, in coordination with some causal sensitivity model, producing tight causal outcome intervals.
We term this approach for \textbf{caus}al outcome intervals via \textbf{mod}ulated \textbf{ens}embles \emph{``Caus-Modens.''}
The idea is to min/max an ensemble's conditional quantiles by reweighing the predictors, yielding individual causal outcome intervals.
First we list the fundamental assumptions for our causal inference.

\begin{assumption}[Potential Outcomes]\label{ass:causal}
  We adopt \citet{rubin}'s first two assumptions for potential outcomes. First, observation tuples of (outcome, assigned treatment, covariates) denoted as $\{(y^{(i)},t^{(i)},x^{(i)})\}_{i=1}^n,$ are \emph{i.i.d} from a single joint distribution. This subsumes the Stable Unit Treatment Value Assumption (SUTVA), where units/individuals cannot depend on one another.
  Secondly, all treatment values have a nonzero chance of assignment for every individual in the data.
\end{assumption}

For a family of outcome predictor models $\mathcal{M}$, we use $p_\mathcal{M}$ to denote probability density functions constrained by one or more models in $\mathcal{M}$---that is, $\mathcal{M}$ conveys the hard constraints implied by the choice of parametrization $\theta$ and any other structural assumption. These models predict an outcome $Y$ due to treatment assignment $T$ and covariate $X$. In Bayesian notation the posterior $P(\Theta\mid\mathcal{D})$, given a dataset $\mathcal{D}$, induces a \emph{posterior predictive} outcome distribution, which is described by a conditional expectation that averages the individual model predictions $p_\mathcal{M}(y\mid t,x;\theta)$:
\begin{equation}\label{eq:predictive}
  p_\mathcal{M}(y\mid t,x; \mathcal{D}) = \E_\Theta\!\big[p_\mathcal{M}(y\mid t,x;\Theta) \mid \mathcal{D}\big].
\end{equation}
In practice, this integration over viable parameters is simulated by Monte Carlo with an ensemble of learned models. For our purposes, $\{\theta^{(j)}\}_{j=1}^m$ is assumed to be i.i.d from an estimator distribution (in the frequentist case) or posterior \citep[in the Bayesian case,][]{li23} with a density denoted as $p(\theta|\mathcal{D})$ in either case.
Our sensitivity analysis requires the estimation of a nominal propensity function as well, denoted by $e_t(x)$, which can be a discrete probability or a continuous density. 

The third potential-outcomes assumption is ignorability: absence of hidden confounders, where 
${{\{(Y_t)_{t\in\mathcal{T}} \indep T\} \mid X}}$.
It states that while the outcome would depend on the assigned treatment, a \emph{potential outcome} for any treatment should not be affected by the treatment assignment, after conditioning on covariates. Our setting allows a bounded violation to the ignorability assumption.

\begin{definition}[Hidden Confounding]\label{def:sens}
  ${{\{(Y_t)_{t\in\mathcal{T}} \notindep T\} \mid X}}$, hence $P(Y_t\mid T,X)$ may differ from $P(Y_t\mid X)$ for the potential outcomes $Y_t$ outside the assigned treatment ($T\neq t$), and similarly the complete propensity $P(T\mid X,Y_t)$ is not the nominal propensity $P(T\mid X)$ for any $Y_t$.
\end{definition}

Whichever sensitivity model is invoked to bound the extent of hidden confounding, all that is required for Caus-Modens is a pair of weight-bounding functions $\underline\omega(t,x), \overline\omega(t,x)$ that are partial identifiers of the potential-outcome probability density function, $p(y_t|x)$. We introduce one layer of indirection by referring to potential \emph{outcome models} $\theta_t$, heterogeneous in treatment $t$ and covariate $x$ (conditioning on the latter,) that can only be partially identified by means of the learned outcome model $\theta$. The real potential outcomes are therefore (partially) identified by marginalization over the potential models: $p(y_t|x) = \int p(y|t,x;\theta_t)\, p(\theta_t|x;\mathcal{D}) \diff\theta_t$, assuming integrability. The role of the weights is in the relation $p(\theta_t|x;\mathcal{D})=\omega(\theta,t,x)p(\theta|t,x;\mathcal{D}),$ where $p(\theta|t,x;\mathcal{D})=p(\theta|\mathcal{D})$ because the learned model is invariant. As mentioned, the weights can only be partially identified by the given sensitivity model. The reason for pushing our causal sensitivity analysis to the level of the outcome \emph{model} is that it can be empirically favorable while remaining largely intuitive.

\begin{assumption}[Sensitivity Model as Weights]\label{ass:weigh}
  The sensitivity model under consideration uses the propensity $e_t(x)$ to produce bounds $0<\underline{\omega}(t,x)\leq 1 \leq \overline{\omega}(t,x)<+\infty$ on weights for partial identification of the outcome model, in the sense that there exists some $\theta\mapsto \omega(\theta,t,x)\in[\underline{\omega}(t,x), \overline{\omega}(t,x)]$ that recovers the true potential outcome model density function, $p(\theta_t|x;\mathcal{D})=\omega(\theta,t,x)\,p(\theta|\mathcal{D})$.
\end{assumption}
This formulation readily accommodates a variety of recently proposed sensitivity models once we pose the \emph{complete propensity} in terms of potential outcome models rather than direct potential outcomes. Instead of $P(T\mid Y_t, X)$, we consider $P(T\mid \Theta_t, X;\mathcal{D})$.
\begin{example}
  For binary treatments, an MSM with violation-of-ignorability $\Gamma>1$ bounds the ratio of complete-propensity odds and nominal-propensity odds to $[1/\Gamma, \Gamma]$, implying~\citep{kallus19} 
  \begin{equation}\label{eq:msm}
    \underline{\omega}(t,x)=e_t(x) + \Gamma[1-e_t(x)], \quad \overline{\omega}(t,x)=e_t(x) + (1/\Gamma)[1-e_t(x)],
  \end{equation}
  where $e_t(x)$ is the nominal propensity of binary treatment $t\in\{0,1\}$. Suppose the complete propensity is denoted as $e_t(x,\theta)$ and gives the propensity of (binary) treatment given that the potential outcome model $\Theta_t$ is known to be $\theta$. The MSM can help partially identify this quantity, which on its own would make it possible to identify the potential outcome model. By the MSM,
  \begin{equation*}
    \Big[\frac{e_t(x)}{1-e_t(x)}\Big]^{-1}\frac{e_t(x,\theta)}{1-e_t(x,\theta)} \in [\Gamma^{-1}, \Gamma^{+1}].
  \end{equation*}
  The bounded ratio of odds permits characterization of the counterfactual in the following equation:
  \begin{align*}
    p(\theta_t|x) &= \underbrace{p(\theta_t\mid T=t, X=x)}_\text{(factual)}e_t(x) + \underbrace{p(\theta_t\mid T=1-t, X=x)}_\text{(counterfactual)}[1-e_t(x)]\\
    &= p(\theta|t, x)e_t(x) + [1-e_t(x,\theta)]p(\theta_t|x),
    \qquad \therefore\quad
    p(\theta_t|x) = \frac{e_t(x)}{e_t(x,\theta)}p(\theta| t, x),
  \end{align*}
  and $\omega(\theta,t,x) \triangleq e_t(x)/e_t(x,\theta)$ according to Assumption~\ref{ass:weigh}, partially identifiable by Equation~\ref{eq:msm}.
\end{example}
\begin{example}
  For continuous-valued treatments, a $\delta$MSM~\citep{marmarelis22} with parameter $\Gamma>1$ likewise defines $[\underline{\omega}, \overline{\omega}]$ in terms of analytically tractable integrals over the propensity density.
\end{example}

Formally, Assumption~\ref{ass:weigh} can be derived as a consequence of using a sensitivity model on the Radon-Nikodym derivative of the potential model $\Theta_t$ with respect to the learned model $\Theta$, while assuming absolute continuity between the two distributions:
\begin{equation*}
  \omega(\theta,t,x) \triangleq \frac{\diff P(\Theta_t = \theta\mid X=x; \mathcal{D})}{\diff P(\Theta = \theta \mid T=t, X=x; \mathcal{D})} = \frac{p(\theta_t \mid x;\mathcal{D})}{p(\theta\mid\mathcal{D})}.
\end{equation*}
Moving forward, we denote empirical propensity estimates as $\tilde e_t(x)$ and hence the empirical weights as $\tilde\omega(\theta,t,x)$ with sensitivity bounds $[\underline{\tilde{\omega}}, \overline{\tilde{\omega}}]$. These are learned from the training data with sample size $n$. On the other hand, posterior quantities approximated by a finite ensemble of size $m$ shall use the hat symbol. The finite-sample version of Equation~\ref{eq:predictive-potential} below, for instance, is denoted as $\hat p_\mathcal{M}(y_t\mid x;\mathcal{D})$.

\subsection{Sensitivity Analysis on Quantiles via Ensemble}\label{sec:quant}

Our proposal is to apply the weights from Assumption~\ref{ass:weigh} directly over the outcome models. Concretely, the potential outcomes are described as a weight-modulated version of Equation~\ref{eq:predictive}:
\begin{equation}\label{eq:predictive-potential}
  p_\mathcal{M}(y_t\mid x; \mathcal{D}) = \E_\Theta\!\big[\omega(\Theta,t,x)\ p_\mathcal{M}(y\mid t,x;\Theta) \mid \mathcal{D}\big], \quad \text{where } \omega(\cdot|t,x)\in[\underline\omega(t,x), \overline\omega(t,x)].
\end{equation}

Prediction intervals for individual outcomes would take the form $[ {F_\omega^{-1}(\alpha/2)},\ {F_\omega^{-1}(1-\alpha/2)} ]$ 
with expected miscoverage $\alpha$ (mirroring the conformal usage,) where $F_\omega(y)$ is the cumulative density of $p_\mathcal{M}(y_t\mid x; \mathcal{D})$, given in Equation~\ref{eq:predictive-potential}. Partial identification entails an optimization over the outcome interval for maximal ignorance as admitted by the sensitivity model. We construct the program
\begin{equation}\label{eq:intervals-minimax}
  \begin{aligned}
  Y_t\mid X \ \in &\Big[ \inf F_{\omega_1}^{-1}(\alpha/2),\  \ \sup F_{\omega_2}^{-1}(1-\alpha/2) \Big],\\
  \text{subject to}\quad &\omega_1(y),\ \omega_2(y) \in \big[\underline\omega(t,x),\ \overline\omega(t,x)\big],\\
  &\text{$F_{\omega_1}(y)$ and $F_{\omega_2}(y)$ are probability distributions.} 
  \end{aligned}
\end{equation}

A globally optimal greedy solution to the finite-sample problem with an ensemble is presented in Supplementary Algorithm~\ref{alg:minimax}, with the optimality criterion addressed in Theorem~\ref{thm:opt}. 
\section{Estimation Properties}\label{sec:theory}

Our main assumption beyond Assumptions~\ref{ass:causal}~\&~\ref{ass:weigh} that enables a simple coverage guarantee of causal outcomes $Y_t$ is that they are independently generated by some unobserved $\Theta_t\sim P(\Theta_t|\mathcal{D})$. This requirement, marked by the subscript $\mathcal{M}$, aligns with our parametric setting and a Bayesian perspective. However, we acknowledge that this result is not as general as the conformal alternatives. We note, additionally, that the empirical evaluations of \S\ref{sec:bench} do not necessarily enforce these conditions.


\begin{lemma}[Empirical Coverage]\label{lem:finite-sample} 
  For fixed values $t$, $x$, and $\alpha\in(0,1)$, consider empirical weights $\tilde\omega(\theta,t,x)$. Let $\hat F_{\tilde\omega}$ be the cumulative distribution of the empirical, finite-ensemble estimate for the potential outcome of Equation~\ref{eq:predictive-potential}, i.e.~$\hat p_\mathcal{M}(y_t\mid x; \mathcal{D}) \,=\, \hat\E_m[\,\tilde\omega(\Theta,t,x)\times p_\mathcal{M}(y\mid t,x;\Theta) \mid \mathcal{D}\,]$. Then for any $\varepsilon>0$ and $\beta=\alpha+\varepsilon+2\E\abs{\tilde\omega - \omega}$, it holds with probability at least $1-\beta$ that
  \begin{equation*}
    \mathbb{P}_\mathcal{M}\big[Y_t \in \big( \hat F_{\tilde\omega}^{-1}(\alpha/2),\  \hat F_{\tilde\omega}^{-1}(1-\alpha/2)\big) \bigm| X=x\big] \ >\  1-2\exp{-m\varepsilon^2/2}.
  \end{equation*}
\end{lemma}

We blend the finite-sample coverage result in Lemma~\ref{lem:finite-sample} with partial identification. Theorem~\ref{thm:valid} characterizes the validity of the causal-outcome intervals from a finite ensemble of size $m$.

\begin{theorem}[Valid Partial Identification]\label{thm:valid}
  For fixed values $t$, $x$, and $\alpha\in(0,1)$, consider weight boundary estimates $[\underline{\tilde{\omega}}, \overline{\tilde{\omega}}]$ yielded from a sensitivity model according to Assumption~\ref{ass:weigh}. Estimating a solution to the program in Equation~\ref{eq:intervals-minimax} produces outcome intervals with hidden-confounding ignorance. Assume for the admitted extrema $\inf\hat F_{\tilde\omega}^{-1}(\alpha/2)$ and $\sup\hat F_{\tilde\omega}^{-1}(1-\alpha/2)$, that $\tilde\omega(\Theta) \in \{\underline{\tilde\omega}, \overline{\tilde\omega}\}$ almost surely. Now let $\beta=\alpha+\varepsilon+2\E[\,\abs{\underline{\tilde\omega} - \underline{\omega}} \vee \abs{\overline{\tilde\omega} - \overline{\omega}}\,]$ for any margin constant $\varepsilon>0$. In this case, with probability at least $1-\beta$,
  \begin{equation}\label{eq:coverage}
    \mathbb{P}_\mathcal{M}\big[Y_t \in \big( \inf\hat F_{\tilde\omega}^{-1}(\alpha/2),\  \sup\hat F_{\tilde\omega}^{-1}(1-\alpha/2)\big) \bigm| X=x\big] \ >\  1-2\exp{-m\varepsilon^2/2}.
  \end{equation}
\end{theorem}

Next, we justify our Supplementary Algorithm~\ref{alg:minimax} by revealing a global optimality condition that can be reached greedily. Theorem~\ref{thm:opt} suggests a simple, monotonically nondecreasing update rule for an optimization algorithm: find pairs of ensemble components that disprove the optimality condition, and transfer weight between them. 

\begin{theorem}[Global Optimality Condition]\label{thm:opt} 
  The weight assignments $\omega(\theta^{(i)}) \in [\underline\omega, \overline\omega]$ for a predictor ensemble $\{\theta^{(i)}\}_{i=1}^m$ maximize the $\beta$-quantile of the finite weighted mixture in the space of all admissible weight assignments \underline{if and only if} there exists no pair of mixture components $(\theta^{(j)}, \theta^{(k)})$ such that weight can be transferred from $j$ to $k$, i.e.\ $\omega(\theta^{(j)}) > \underline\omega$ and $\omega(\theta^{(k)}) < \overline\omega$, and $j$ has more leftward mass than $k$, i.e.\ $F(q;\theta^{(j)}) > F(q;\theta^{(k)})$ where $q$ is the current $\beta$-quantile: $\beta = m^{-1}\sum_i \omega(\theta^{(i)}) F(q; \theta^{(i)})$.
\end{theorem}


Some of the empirical tightness of the outcome intervals might stem from the preservation of continuity in the partially identified densities; by the definition of Lipschitz continuity on the reals, 
\begin{proposition}[Continuity of Outcome Density]
  If the predictor densities $p_\mathcal{M}(y\mid t,x;\theta_i)$ are $C$-Lipschitz continuous, then the posterior outcome density $p_\mathcal{M}(y_t\mid x; \mathcal{D})$ is $\overline\omega C$-Lipschitz.
\end{proposition}

On the other hand, a sample-based reweighing scheme like from \citet{kallus19,jesson21} does not preserve any implied continuity of the partially identified probability density.
\FloatBarrier

\section{Empirical Evaluations}\label{sec:bench}

We present three benchmarks comparing the tightness of the outcome intervals produced by Caus-Modens and the prevailing conformalized causal sensitivity analyses. As discussed in \S\ref{sec:intro}, these conformal approaches encompass the state of the art in partial idenfiticaiton of individual outcomes and not outcome expectations. We first list the baselines and then detail the evaluation procedure.

\paragraph{The baseline methods.}
We consider various combinations and ablations of conformal sensitivity analysis (CSA)~\citep{yin22,jin23} with the two state-of-the-art conformal backbones: distributional conformal prediction (DCP)~\citep{chernozhukov21} and conformalized quantile regression (CQR)~\citep{romano19}. The CSA studies relied on CQR for their implementations. In the meanwhile an even more adaptive procedure, DCP, was proposed. For completeness in our analysis we constructed a ``supercharged'' baseline that combined CSA with DCP. 

Since we learned an entire ensemble for each benchmark, we usually allowed the conformal alternatives to also harness the empirical uncertainties captured by the ensemble. Again, this was done in an attempt to be as favorable to the conformal alternatives as possible. Ensembled baselines were marked by the ``Ens-'' prefix, and the one baseline without that simply used a single model drawn from the ensemble.
The predictive modeling foundations for all methods were kept the same so that there was no question about differential modeling performance leading to different results between Caus-Modens and baselines. This ensured the benchmarks were commensurate. For instance, whereas CQR normally calls for quantile regression, we fed it quantiles of the ensemble-marginalized distributional prediction. We list the actual baselines implemented:
\begin{itemize} \itemsep 0em
  \item Ens-CSA-DCP --- the main conformal baseline with all the beneficial components;
  \item Ens-CSA-CQR --- similar to the above, but with the more standard CQR;
  \item CSA-DCP --- the non-ensembled ablation;
  \item Ens-DCP --- the non-CSA (non-causal) ablation.
\end{itemize}

\paragraph{How tightness is measured.}
Most of our results were reported with a concept of coverage efficiency. As is customary with studies on conformal inference, we set a target coverage level. Then we observed the size of the intervals required to achieve that level of coverage on a causal test set where treatments were de-confounded (the smaller the intervals, the more \emph{efficient} the coverage.) The logic of this strategy is that a tighter partial identification should require less implied hidden confounding (via the sensivitiy model) to cover the causal outcomes, relative to a more conservative method that does not utilize the sample statistics or problem assumptions effectively. Each result section (\S\ref{sec:bench-ihdp}, \S\ref{sec:bench-pbmc}, \S\ref{sec:bench-aita}) defined a domain-specific cost function to measure the size of the outcome intervals. Tighter intervals had lower cost. We always used \citet{tan}'s sensitivity model for binary treatments (MSM) and varied its single parameter $\Gamma$ for the extent of violation to the ignorability assumption. We explored the landscape between and including $\Gamma=$~1, where ignorability holds, and $\Gamma=$~50, where all methods would plateau. We used binary search to identify the smallest $\Gamma\coloneqq\Gamma^*\in[1,50]$ that achieves a target coverage, like 95\% of the test set. That point could be $\Gamma=$~1, the no-hidden-confounding condition. On the other hand, an experiment was classified as a failure if the method never reached the target coverage. The cost function evaluated at a successful $\Gamma^*$ was termed the \emph{coverage cost}.

\begin{figure}[ht]
  \begin{minipage}{0.48\linewidth}
    \scalebox{0.79}{ 
      \input{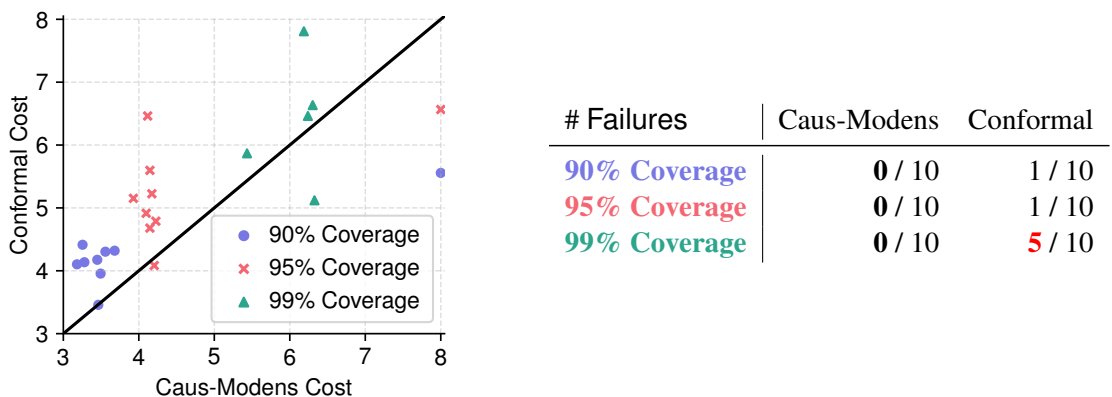}}
  \end{minipage}
  \hfill
  \begin{minipage}{0.5\linewidth}
    \begin{tabular}{l | r @{\hspace{1em}} r }
      \# \textsf{Failures} & \text{Caus-Modens} & \text{Conformal} \\ 
      \midrule
      \textcolor{nice-blue}{\bf 90\% Coverage} & \textbf{0} / 10 & 1 / 10 \\
      \textcolor{nice-red}{\bf 95\% Coverage} & \textbf{0} / 10 & 1 / 10 \\
      \textcolor{nice-green}{\bf 99\% Coverage} & \textbf{0} / 10 & \textcolor{red!!black}{\bf 5} / 10 \\
    \end{tabular}\vspace{1.5em}
    
  \end{minipage}\vspace{-0.75em}
  \caption{\label{fig:ihdp-scatter}Coverage costs of Caus-Modens versus the main ensemble-conformal baseline, Ens-CSA-DCP, for the ten IHDP realizations used by~\citet{louizos}. In the scatter plot, we only display the cost pairs, clipped at 8.0, where both methods achieved the target coverage. The table shows that all the failures to reach adequate coverage occurred on the conformal method. } 
\end{figure}

\paragraph{Training the predictors.}
Whereas Caus-Modens was conceived in a Bayesian framework, in practice, deep ensembles tend to achieve better accuracy than Bayesian neural networks and similarly quantify empirical uncertainty~\citep{lakshminarayanan17,pearce18,fort19,rahaman21}. Caus-Modens ultimately requires a sample of predictive models, whether from a posterior or an estimator distribution. The focus of our evaluations is the sensitivity analyses once models have been learned. Hence, we use deep ensembles of cardinality 16 in our reported benchmarks. 
We trained fully connected feedforward neural networks with sigmoid activations for both the outcome and the propensity predictors. Hyperparameter and architecture selection was done by grid search. Ensembles were trained by maximum likelihood on bootstrap-resampled training sets and randomly initialized weights. Caus-Modens and all the baselines relied on this set of predictors, either in whole or by randomly drawing individual models in the case of non-ensemble ablations.

\subsection{Classical Benchmark \texttt{(IHDP)}}\label{sec:bench-ihdp}


The \texttt{IHDP} dataset in causal literature is a semi-synthetic classical benchmark for CATE estimation~\citep[e.g.][]{louizos,jesson21-bald,samothrakis22}. It contains binary treatments with covariate shift, and simulated real-valued outcomes, for 747 individuals. The original covariates are eight real and nineteen binary attributes. To induce hidden confounding, we obscured the binary covariates. The benchmark task was to predict the $T=1$ (potential) outcomes of the test set.
Due to the smallness of the sample, we randomly allocated 10\% of the data to the validation set and 20\% to the test set. In addition to hyperparameter selection, the validation set served for calibration in the conformal baselines for maximal resourcefulness. The whole calibration set consisted of 3/7$^\text{ths}$ of the otherwise-labeled training set for an ultimate 50-50 estimation-calibration split, as is recommended with split conformal prediction (SCP)~\citep{papadopoulos02}. In other words, Caus-Modens utilized the entire training set, and the conformal baselines used 4/7$^\text{ths}$ of the training set for estimation and the rest for calibration. We also tested a 75-25 SCP split rather than 50-50, to similar effect. 

\paragraph{Cost function \& results.}
The cost function was the absolute length of the interval scaled to the empirical standard deviation of the outcomes.
We first tested Caus-Modens against Ens-CSA-DCP for three target coverages shown in Figure~\ref{fig:ihdp-scatter}. Caus-Modens produced tighter intervals with Wilcoxon signed-rank test $p<0.05$. This dataset was noteworthy for the occurrence of failures in the conformal approach and complete success in Caus-Modens for achieving the target coverage. Supplemental Table~\ref{tab:ihdp-splits} shows how other conformal configurations induced more failures. For Caus-Modens we found that the size of the ensemble beyond 16 predictors ceased to impact the coverage cost.

\subsection{Novel Semi-synthetic Benchmark \texttt{(PBMC)}}\label{sec:bench-pbmc}
Recent widely celebrated single-cell RNA sequencing (scRNAseq) modalities have enabled an unprecedented view into human physiology~\citep{jovic22}. The complex relations between the expressions of roughly 20,000 genes makes it a good source for benchmark datasets with unintuitive statistics. We obtained a relatively clean dataset of well-characterized peripheral blood mononuclear cells (PBMCs)~\citep{kang} and randomly projected the gene expressions into 32 observed and 32 unobserved confounders, as well as a treatment variable that was discretized to binary values. The simulated outcome was a completely random quadratic form of all these 32+32+1 variables, ensuring arbitrary relations between treatment assignment, confounders, and outcome.

A current shortcoming of partial identifiers for expectations of causal quantities, like ATE and CATE, is that they were not designed for heavy-tailed outcomes. We showcase Caus-Modens in this light, using the Cauchy distribution for simulated \texttt{PBMC} outcomes. The Cauchy distribution has several scientific uses, including the modeling of physical \& financial phenomena~\citep[e.g.][]{kagan94} and specifying priors for variance~\citep{gelman06}.
It is considered ``pathological'' because it has no mean or higher moments. The sample mean is also Cauchy distributed---for it is a stable distribution~\citep[][]{nolan20}---and diverges in large samples. However, a viable alternative is to estimate the tail parameters by maximum likelihood~\citep[e.g.][]{huisman01, taleb20}.
Parametric approaches are paramount to characterizing pathological distributions like the Cauchy, which are punctuated by extreme rare events.
This simple benchmark highlights the value of inductive bias.

\paragraph{Cost function \& results.}
For an interpretable measure than can be aggregated across multiple experiments, the cost function for Cauchy-outcome intervals was the interval length scaled to the smallest achieved length in each setting.
Table~\ref{tab:pbmc} displays these relative costs for a high coverage target of 99\%, evaluating each method's ability to characterize Cauchy tails. Caus-Modens achieved \emph{significantly} lower costs than the CSA baselines while meeting target coverage on average, and the non-causal Ens-DCP had similar cost for greater miscoverage, with failure on average. 

\begin{table}[ht]\centering 
  \begin{tabular}{l | r r r } 
    Method & Achieved Coverage $\uparrow$ & Coverage Cost $\downarrow$ & Avg.\ Coverage Loss $\downarrow$ \\
    \midrule
    Caus-Modens & \textbf{99.15} (0.20) \% & \textbf{0.28} (0.15) & 0.028 \% pts \\
    Ens-CSA-DCP & \textbf{99.58} (0.32) \% & 1.51 (1.94)          & 0.002 \% pts \\
    Ens-CSA-CQR & \textbf{99.57} (0.32) \% & 1.51 (1.84)          & 0.003 \% pts \\
    CSA-DCP     & \textbf{99.60} (0.32) \% & 1.50 (1.82)          & 0.002 \% pts \\
    Ens-DCP     & 98.95 (0.45) \%          & 0.30 (0.20)          & 0.206 \% pts \\
  \end{tabular}\vspace{0.50em}

  \caption{\label{tab:pbmc}\texttt{PBMC} results from 16 independent dataset generations and inferences. We set the random seed to the predetermined value 0 prior to generation for reproducibility and transparency. We present average achieved coverages and standard deviations for a target of 99\%, accompanied by relative coverage costs for the trials that met the target, and the average nonnegative loss in coverage percentage points, which was positive for trials with coverage below 99\%. }
\end{table}

\subsection{Novel Benchmark via GPT-4 \texttt{(AITA)}}\label{sec:bench-aita}

Semi-synthetic causal benchmarks like \texttt{PBMC} can be designed to harness the arbitrary statistical relations in real data. Still, the outcome must have pre-specified functional relations with the treatment and confounders.
With the proliferation of causal-infernece studies proposing new methods for various settings, there is a need for flexible yet realistic benchmarks. In this result section we took a step in building a new kind of observational dataset that includes intervention results without the challenges of actually bringing in a randomized control experiment for testing the causal inference.
We used the celebrated large language model (LLM) GPT-4~\citep{openai23} that has demonstrated remarkable capabilities in emulating human text~\citep{brown20}. 
One can use an LLM to sample complex outcomes from observational datasets and also query textual \emph{interventions}. We seek to promote this usage of large generative models for benchmarking causal inference~\citep{curth}.

We framed the novel inference task in the format of the r/AmITheAsshole subreddit (hence the name of this benchmark, \texttt{AITA}.) The subreddit is a scientifically attractive setting because the rules and structure of the forum are clean: users post personal stories of conflict, and comments offer opinions on whether the author was at fault in the way the story transpired. A verdict is determined by the upvote mechanism on comments. Data from this subreddit have recently served as a vessel for human perspectives~\citep{botzer22} and moral judgment~\citep{plepi22} in the computational social sciences.
For the sake of a causal benchmark, we asked GPT-4 to act as moral arbiter on real posts from the subreddit~\citep{obrien20}. That way there would be no doubt about the real-world salience of the data, while permitting interventions via the LLM. The treatment variable was the customary self-identified gender, which is binary between `F' and `M' (a limiting and problematic format.) Nevertheless, this variable allowed us to assess a bias in GPT-4's verdicts.

The mechanics of GPT-4's ``intuitive process'' are so complex that it would be difficult to predict its moral judgments through a much simpler outcome predictor that would necessarily be trained on a relatively small sample of text embeddings: the \texttt{AITA} posts that have discernible gender markers. We changed this benchmark in order to simplify the prediction problem. Concretely, the outcome predictor was tasked with denoising an artificially noised GPT-4 verdict, utilizing the gender (treatment) and topic (covariates) information of the post. See Supplemental Figure~\ref{fig:aita} for a diagram. Topics were represented by five-dimensional embeddings like in BERTopic~\citep{grootendorst22}. To strengthen the bias in GPT-4, we coupled the real indicated gender with synthetic ages so that authors were either 30-year old men ($T=0$) or 70-year old women ($T=1$). Caus-Modens and baselines were tasked with predicting the causal effect of $T$ on the denoised verdict.

\begin{figure}[ht]\centering
  \scalebox{0.80}{ 
    \input{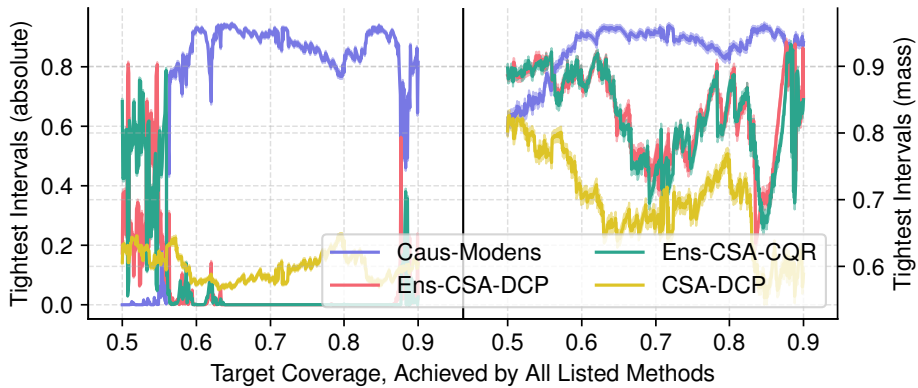}}\vspace{-0.5em}\\
  \caption{\label{fig:aita-shares}Share of the test set that each method produced the tightest intervals, shown in absolute (left) and mass (right) units, at a wide range of target coverages. Lines are widened by standard errors. The four listed methods always achieved the target, whereas the unlisted Ens-DCP failed frequently. The presence of ties, particularly in the right subplot, allows the shares to add up to more than unit.}
\end{figure}

\paragraph{Cost function \& results.}
The GPT-4 verdict was a number from 1--99, afterwards rescaled to the unit interval, and then logit-transformed to be modeled by normal distributions. These outcomes had a standard deviation of $0.62$ and the artificial noise standard deviation was selected to be $0.5$. After learning the outcome (and propensity,) we focused on the woman gender arm of potential outcomes for the intervention testing set. Verdict-noise variance out of sample was reduced by 40\% with the predictors. We chose two cost functions for evaluating coverage efficiency: the absolute units of outcome interval size, and \emph{mass} units computed as the integral of the empirical marginal outcome distribution along the interval. Figure~\ref{fig:aita-shares} shows the rate of tightest intervals for all the evaluated methods that reached sufficient coverage, in both units and at a wide array of coverage targets.

At 70\% coverage, the verdict overlap between genders is smaller with Caus-Modens outcome intervals than with Ens-CSA-DCP for 20.0\,(0.8)\,\% of the posts, with the rest being equal in the empirical probability mass units, and none in the other direction. This suggests that Caus-Modens identifies more gender bias in GPT-4's moral judgments.
The doubly robust and debiased ATE estimator~\citep{yao} for gender effect suggests that $\text{ATE} = -0.07$ with Student $t$-test $p<0.05$, likewise revealing a gender bias towards young men being more wrong than old women according to GPT-4. Table~\ref{tab:aita-confusion} compares GPT-4 verdicts to the original Reddit verdicts and displays coverage costs for different judgment regimes. Caus-Modens consistently outperformed Ens-CSA-DCP, which was the best of the baselines according to Figure~\ref{fig:aita-shares}, even in empirical mass units.

\begin{table}[ht]
  \begin{tabular}{l | r l }
    \textbf{Counts} & Reddit right & Reddit wrong \\
    \midrule
    GPT-4 right & 1303 [58\%] & 276 [12\%] \\
    GPT-4 wrong & 430 [19\%] & 250 [11\%] \\
  \end{tabular}
  \hfill
  \begin{tabular}{l | r l }
    \textbf{Costs} & Reddit right & Reddit wrong \\
    \midrule
    GPT-4 right & \textbf{0.36}* / 0.38 & \textbf{0.38} / 0.39 \\ 
    GPT-4 wrong & \textbf{0.39}* / 0.41 & 0.35 / 0.37 \\
  \end{tabular} 
  \vspace{0.50em}
  \caption{\label{tab:aita-confusion}Confusion matrix of Reddit versus GPT-4 verdicts (\textbf{left}) and the 70\%-coverage costs of Caus-Modens against Ens-CSA-DCP for those stratified posts (\textbf{right}). Bold: $p<0.05$; asterisk: $p<0.01$. As the GPT-4 verdict was given on a sliding scale, we chose a right/wrong threshold by equalizing its marginal rate with Reddit's. If we chose the midpoint of the verdict spectrum, there would have been many more wrong than right verdicts from GPT-4. This can be explained by a higher threshold for the designation of ``asshole'' to the author of a post, as is the Reddit protocol. }
\end{table}

\section{Discussion}\label{sec:disc}


We found that even the state of the art in adaptive conformal prediction with causal sensitivity analysis is overly conservative in our three benchmarks, \texttt{IHDP}, \texttt{PBMC}, and \texttt{AITA}. In \texttt{IHDP} (\S\ref{sec:bench-ihdp}), the conformal baselines surprisingly failed at least once out of ten trials and the costs of non-failures tended to be larger than Caus-Modens' at the same target coverage. Caus-Modens did not fail at all for \texttt{IHDP}. The failure rate of the conformalized sensitivity analyses highlights real-world pitfalls of finite-sample statistical guarantees that conformal inference enjoys. For instance, validity actually depends on correct specification of the propensity~\citep[see Lemma~3,][]{yin22}.

In the much larger \texttt{PBMC} (\S\ref{sec:bench-pbmc}) benchmark, Caus-Modens achieved the tightest coverage for the majority of the trials. \texttt{PBMC} leveraged the relations between gene expressions in human cells by randomly projecting them into confounders and treatment. Further, it enabled the demonstration of Caus-Modens quantile predictions for extremely heavy tails. In the novel \texttt{AITA} benchmark (\S\ref{sec:bench-aita}), which used an LLM to generate extremely complex causal outcomes with access to interventions, we also demonstrated that the outcome intervals produced by Caus-Modens were markedly more informative than the intervals produced by the conformal baselines.

\paragraph{Societal concerns.}
We explicitly assert that GPT-4's verdicts on morality like in the \texttt{AITA} scheme must not be confused with anything approximating proper moral judgments. The verdicts must be viewed as a phenomenon detached from the philosophy of morality and ethics. Further, the biases isolated in these contrived experiments do not necessarily reflect the general biases present in GPT-4.

\paragraph{Limitations.}
As with any causal inference on observational studies guided by a sensitivity model, care must be taken to not overly trust the assumptions surrounding the data-generating process (namely on constraints for hidden confounders) nor the estimates. Our method elevates the role of empirical uncertainty relative to many prior works. However, in contrast with the conformal alternatives, Caus-Modens may be less conservative by relying more on parametric, structural, and inductive constraints. Hence, it could also be more vulnerable to model misspecification.

\paragraph{Conclusion.}
Our simple ensemble-based partial identification of outcome quantiles is a promising approach to prediction intervals that leverages the inductive biases of deep models. In addition to coverage efficiency, it accommodates various sensitivity models and adapts them to a novel formulation of potential posteriors that justifies the weight-modulation of an ensemble. We recommend investigating how to regularize Caus-Modens to be sharper in its partial identification, especially in general (non-binary) treatment domains that are less explored.

\FloatBarrier


\bibliography{refs}

\begin{thebibliography}{55}
\providecommand{\natexlab}[1]{#1}
\providecommand{\url}[1]{\texttt{#1}}
\expandafter\ifx\csname urlstyle\endcsname\relax
  \providecommand{\doi}[1]{doi: #1}\else
  \providecommand{\doi}{doi: \begingroup \urlstyle{rm}\Url}\fi

\bibitem[Athey et~al.(2019)Athey, Tibshirani, and Wager]{athey19}
Susan Athey, Julie Tibshirani, and Stefan Wager.
\newblock Generalized random forests.
\newblock \emph{The Annals of Statistics}, 47\penalty0 (2):\penalty0 1148--1178, 2019.

\bibitem[Bica et~al.(2020)Bica, Jordon, and van~der Schaar]{bica20}
Ioana Bica, James Jordon, and Mihaela van~der Schaar.
\newblock Estimating the effects of continuous-valued interventions using generative adversarial networks.
\newblock \emph{Advances in Neural Information Processing Systems}, 33:\penalty0 16434--16445, 2020.

\bibitem[Botzer et~al.(2022)Botzer, Gu, and Weninger]{botzer22}
Nicholas Botzer, Shawn Gu, and Tim Weninger.
\newblock Analysis of moral judgment on reddit.
\newblock \emph{IEEE Transactions on Computational Social Systems}, 2022.

\bibitem[Brown et~al.(2020)Brown, Mann, Ryder, Subbiah, Kaplan, Dhariwal, Neelakantan, Shyam, Sastry, Askell, et~al.]{brown20}
Tom Brown, Benjamin Mann, Nick Ryder, Melanie Subbiah, Jared~D Kaplan, Prafulla Dhariwal, Arvind Neelakantan, Pranav Shyam, Girish Sastry, Amanda Askell, et~al.
\newblock Language models are few-shot learners.
\newblock \emph{Advances in neural information processing systems}, 33:\penalty0 1877--1901, 2020.

\bibitem[Chernozhukov et~al.(2018)Chernozhukov, Chetverikov, Demirer, Duflo, Hansen, Newey, and Robins]{chernozhukov17}
Victor Chernozhukov, Denis Chetverikov, Mert Demirer, Esther Duflo, Christian Hansen, Whitney Newey, and James Robins.
\newblock {Double/debiased machine learning for treatment and structural parameters}.
\newblock \emph{The Econometrics Journal}, 21\penalty0 (1):\penalty0 C1--C68, 01 2018.

\bibitem[Chernozhukov et~al.(2021)Chernozhukov, W{\"u}thrich, and Zhu]{chernozhukov21}
Victor Chernozhukov, Kaspar W{\"u}thrich, and Yinchu Zhu.
\newblock Distributional conformal prediction.
\newblock \emph{Proceedings of the National Academy of Sciences}, 118\penalty0 (48):\penalty0 e2107794118, 2021.

\bibitem[Curth et~al.(2021)Curth, Svensson, Weatherall, and van~der Schaar]{curth}
Alicia Curth, David Svensson, Jim Weatherall, and Mihaela van~der Schaar.
\newblock Really doing great at estimating cate? a critical look at ml benchmarking practices in treatment effect estimation.
\newblock In \emph{Thirty-fifth conference on neural information processing systems datasets and benchmarks track (round 2)}, 2021.

\bibitem[Dorn and Guo(2022)]{dorn22}
Jacob Dorn and Kevin Guo.
\newblock Sharp sensitivity analysis for inverse propensity weighting via quantile balancing.
\newblock \emph{Journal of the American Statistical Association}, pages 1--13, 2022.

\bibitem[Dorn et~al.(2021)Dorn, Guo, and Kallus]{dorn}
Jacob Dorn, Kevin Guo, and Nathan Kallus.
\newblock Doubly-valid/doubly-sharp sensitivity analysis for causal inference with unmeasured confounding.
\newblock \emph{arXiv preprint arXiv:2112.11449}, 2021.

\bibitem[Edelman et~al.(2022)Edelman, Goel, Kakade, and Zhang]{edelman22}
Benjamin~L Edelman, Surbhi Goel, Sham Kakade, and Cyril Zhang.
\newblock Inductive biases and variable creation in self-attention mechanisms.
\newblock In \emph{International Conference on Machine Learning}, pages 5793--5831. PMLR, 2022.

\bibitem[Fort et~al.(2019)Fort, Hu, and Lakshminarayanan]{fort19}
Stanislav Fort, Huiyi Hu, and Balaji Lakshminarayanan.
\newblock Deep ensembles: A loss landscape perspective.
\newblock \emph{arXiv preprint arXiv:1912.02757}, 2019.

\bibitem[Gelman(2006)]{gelman06}
A~Gelman.
\newblock Prior distributions for variance parameters in hierarchical models (comment on an article by browne and draper).
\newblock \emph{Bayesian Analysis}, 1:\penalty0 515--533, 2006.

\bibitem[Glynn et~al.(1996)]{glynn}
Peter~W Glynn et~al.
\newblock Importance sampling for monte carlo estimation of quantiles.
\newblock In \emph{Mathematical Methods in Stochastic Simulation and Experimental Design: Proceedings of the 2nd St. Petersburg Workshop on Simulation}, pages 180--185. Citeseer, 1996.

\bibitem[Grootendorst(2022)]{grootendorst22}
Maarten Grootendorst.
\newblock Bertopic: Neural topic modeling with a class-based tf-idf procedure.
\newblock \emph{arXiv preprint arXiv:2203.05794}, 2022.

\bibitem[Huisman et~al.(2001)Huisman, Koedijk, Kool, and Palm]{huisman01}
Ronald Huisman, Kees~G Koedijk, Clemens J~M Kool, and Franz Palm.
\newblock Tail-index estimates in small samples.
\newblock \emph{Journal of Business \& Economic Statistics}, 19\penalty0 (2):\penalty0 208--216, 2001.

\bibitem[Jaynes(2003)]{jaynes}
Edwin~T Jaynes.
\newblock \emph{Probability theory: The logic of science}.
\newblock Cambridge university press, 2003.

\bibitem[Jesson et~al.(2020)Jesson, Mindermann, Shalit, and Gal]{jesson20}
Andrew Jesson, S{\"o}ren Mindermann, Uri Shalit, and Yarin Gal.
\newblock Identifying causal-effect inference failure with uncertainty-aware models.
\newblock \emph{Advances in Neural Information Processing Systems}, 33:\penalty0 11637--11649, 2020.

\bibitem[Jesson et~al.(2021{\natexlab{a}})Jesson, Mindermann, Gal, and Shalit]{jesson21}
Andrew Jesson, Sören Mindermann, Yarin Gal, and Uri Shalit.
\newblock Quantifying ignorance in individual-level causal-effect estimates under hidden confounding.
\newblock \emph{ICML}, 2021{\natexlab{a}}.

\bibitem[Jesson et~al.(2021{\natexlab{b}})Jesson, Tigas, van Amersfoort, Kirsch, Shalit, and Gal]{jesson21-bald}
Andrew Jesson, Panagiotis Tigas, Joost van Amersfoort, Andreas Kirsch, Uri Shalit, and Yarin Gal.
\newblock Causal-bald: Deep bayesian active learning of outcomes to infer treatment-effects from observational data.
\newblock \emph{Advances in Neural Information Processing Systems}, 34:\penalty0 30465--30478, 2021{\natexlab{b}}.

\bibitem[Jesson et~al.(2022)Jesson, Douglas, Manshausen, Solal, Meinshausen, Stier, Gal, and Shalit]{jesson22}
Andrew Jesson, Alyson~Rose Douglas, Peter Manshausen, Ma{\"e}lys Solal, Nicolai Meinshausen, Philip Stier, Yarin Gal, and Uri Shalit.
\newblock Scalable sensitivity and uncertainty analyses for causal-effect estimates of continuous-valued interventions.
\newblock In Alice~H. Oh, Alekh Agarwal, Danielle Belgrave, and Kyunghyun Cho, editors, \emph{Advances in Neural Information Processing Systems}, 2022.

\bibitem[Jin et~al.(2023)Jin, Ren, and Cand{\`e}s]{jin23}
Ying Jin, Zhimei Ren, and Emmanuel~J Cand{\`e}s.
\newblock Sensitivity analysis of individual treatment effects: A robust conformal inference approach.
\newblock \emph{Proceedings of the National Academy of Sciences}, 120\penalty0 (6):\penalty0 e2214889120, 2023.

\bibitem[Jovic et~al.(2022)Jovic, Liang, Zeng, Lin, Xu, and Luo]{jovic22}
Dragomirka Jovic, Xue Liang, Hua Zeng, Lin Lin, Fengping Xu, and Yonglun Luo.
\newblock Single-cell rna sequencing technologies and applications: A brief overview.
\newblock \emph{Clinical and Translational Medicine}, 12\penalty0 (3):\penalty0 e694, 2022.

\bibitem[Kaddour et~al.(2021)Kaddour, Zhu, Liu, Kusner, and Silva]{kaddour}
Jean Kaddour, Yuchen Zhu, Qi~Liu, Matt~J Kusner, and Ricardo Silva.
\newblock Causal effect inference for structured treatments.
\newblock \emph{Advances in Neural Information Processing Systems}, 34:\penalty0 24841--24854, 2021.

\bibitem[Kagan(1994)]{kagan94}
Yan~Y Kagan.
\newblock Observational evidence for earthquakes as a nonlinear dynamic process.
\newblock \emph{Physica D: Nonlinear Phenomena}, 77\penalty0 (1-3):\penalty0 160--192, 1994.

\bibitem[Kallus(2022)]{kallus22}
Nathan Kallus.
\newblock Treatment effect risk: Bounds and inference.
\newblock In \emph{2022 ACM Conference on Fairness, Accountability, and Transparency}, pages 213--213, 2022.

\bibitem[Kallus and Oprescu(2023)]{kallus23}
Nathan Kallus and Miruna Oprescu.
\newblock Robust and agnostic learning of conditional distributional treatment effects.
\newblock In \emph{International Conference on Artificial Intelligence and Statistics}, pages 6037--6060. PMLR, 2023.

\bibitem[Kallus et~al.(2019)Kallus, Mao, and Zhou]{kallus19}
Nathan Kallus, Xiaojie Mao, and Angela Zhou.
\newblock Interval estimation of individual-level causal effects under unobserved confounding.
\newblock In \emph{The 22nd international conference on artificial intelligence and statistics}, pages 2281--2290. PMLR, 2019.

\bibitem[Kang et~al.(2018)Kang, Subramaniam, Targ, Nguyen, Maliskova, McCarthy, Wan, Wong, Byrnes, Lanata, et~al.]{kang}
Hyun~Min Kang, Meena Subramaniam, Sasha Targ, Michelle Nguyen, Lenka Maliskova, Elizabeth McCarthy, Eunice Wan, Simon Wong, Lauren Byrnes, Cristina~M Lanata, et~al.
\newblock Multiplexed droplet single-cell rna-sequencing using natural genetic variation.
\newblock \emph{Nature biotechnology}, 36\penalty0 (1):\penalty0 89--94, 2018.

\bibitem[Lakshminarayanan et~al.(2017)Lakshminarayanan, Pritzel, and Blundell]{lakshminarayanan17}
Balaji Lakshminarayanan, Alexander Pritzel, and Charles Blundell.
\newblock Simple and scalable predictive uncertainty estimation using deep ensembles.
\newblock \emph{Advances in neural information processing systems}, 30, 2017.

\bibitem[Lei and Cand{\`e}s(2021)]{lei21}
Lihua Lei and Emmanuel~J Cand{\`e}s.
\newblock Conformal inference of counterfactuals and individual treatment effects.
\newblock \emph{Journal of the Royal Statistical Society Series B: Statistical Methodology}, 83\penalty0 (5):\penalty0 911--938, 2021.

\bibitem[Li et~al.(2023)Li, Ding, and Mealli]{li23}
Fan Li, Peng Ding, and Fabrizia Mealli.
\newblock Bayesian causal inference: a critical review.
\newblock \emph{Philosophical Transactions of the Royal Society A}, 381\penalty0 (2247):\penalty0 20220153, 2023.

\bibitem[Louizos et~al.(2017)Louizos, Shalit, Mooij, Sontag, Zemel, and Welling]{louizos}
Christos Louizos, Uri Shalit, Joris~M Mooij, David Sontag, Richard Zemel, and Max Welling.
\newblock Causal effect inference with deep latent-variable models.
\newblock \emph{Advances in neural information processing systems}, 30, 2017.

\bibitem[Manski(2003)]{manski}
Charles~F Manski.
\newblock \emph{Partial identification of probability distributions}, volume~5.
\newblock Springer, 2003.

\bibitem[Marmarelis et~al.(2023)Marmarelis, Haddad, Jesson, Jahanshad, Galstyan, and Ver~Steeg]{marmarelis22}
Myrl~G Marmarelis, Elizabeth Haddad, Andrew Jesson, Neda Jahanshad, Aram Galstyan, and Greg Ver~Steeg.
\newblock Partial identification of dose responses with hidden confounders.
\newblock In \emph{Uncertainty in Artificial Intelligence}, pages 1368--1379. PMLR, 2023.

\bibitem[Nie et~al.(2021)Nie, Ye, qiang liu, and Nicolae]{nie}
Lizhen Nie, Mao Ye, qiang liu, and Dan Nicolae.
\newblock {\{}VCN{\}}et and functional targeted regularization for learning causal effects of continuous treatments.
\newblock In \emph{International Conference on Learning Representations}, 2021.

\bibitem[Nolan(2020)]{nolan20}
John~P Nolan.
\newblock Univariate stable distributions.
\newblock \emph{Springer Series in Operations Research and Financial Engineering, DOI}, 10:\penalty0 978--3, 2020.

\bibitem[O'Brien(2020)]{obrien20}
Elle O'Brien.
\newblock {iterative/aita\_dataset: Praw rescrape of entire dataset}, February 2020.
\newblock URL \url{https://doi.org/10.5281/zenodo.3677563}.

\bibitem[OpenAI(2023)]{openai23}
OpenAI.
\newblock Gpt-4 technical report, 2023.

\bibitem[Papadopoulos et~al.(2002)Papadopoulos, Proedrou, Vovk, and Gammerman]{papadopoulos02}
Harris Papadopoulos, Kostas Proedrou, Volodya Vovk, and Alex Gammerman.
\newblock Inductive confidence machines for regression.
\newblock In \emph{Machine Learning: ECML 2002: 13th European Conference on Machine Learning Helsinki, Finland, August 19--23, 2002 Proceedings 13}, pages 345--356. Springer, 2002.

\bibitem[Pearce et~al.(2018)Pearce, Brintrup, Zaki, and Neely]{pearce18}
Tim Pearce, Alexandra Brintrup, Mohamed Zaki, and Andy Neely.
\newblock High-quality prediction intervals for deep learning: A distribution-free, ensembled approach.
\newblock In \emph{International conference on machine learning}, pages 4075--4084. PMLR, 2018.

\bibitem[Plepi et~al.(2022)Plepi, Neuendorf, Flek, and Welch]{plepi22}
Joan Plepi, B{\'e}la Neuendorf, Lucie Flek, and Charles Welch.
\newblock Unifying data perspectivism and personalization: An application to social norms.
\newblock In \emph{Proceedings of the 2022 Conference on Empirical Methods in Natural Language Processing}, pages 7391--7402, Abu Dhabi, United Arab Emirates, December 2022. Association for Computational Linguistics.
\newblock URL \url{https://aclanthology.org/2022.emnlp-main.500}.

\bibitem[Rahaman et~al.(2021)]{rahaman21}
Rahul Rahaman et~al.
\newblock Uncertainty quantification and deep ensembles.
\newblock \emph{Advances in Neural Information Processing Systems}, 34:\penalty0 20063--20075, 2021.

\bibitem[Rame et~al.(2022)Rame, Kirchmeyer, Rahier, Rakotomamonjy, patrick gallinari, and Cord]{rame}
Alexandre Rame, Matthieu Kirchmeyer, Thibaud Rahier, Alain Rakotomamonjy, patrick gallinari, and Matthieu Cord.
\newblock Diverse weight averaging for out-of-distribution generalization.
\newblock In Alice~H. Oh, Alekh Agarwal, Danielle Belgrave, and Kyunghyun Cho, editors, \emph{Advances in Neural Information Processing Systems}, 2022.

\bibitem[Romano et~al.(2019)Romano, Patterson, and Candes]{romano19}
Yaniv Romano, Evan Patterson, and Emmanuel Candes.
\newblock Conformalized quantile regression.
\newblock \emph{Advances in neural information processing systems}, 32, 2019.

\bibitem[Rosenbaum and Rubin(1983)]{rosenbaum83}
Paul~R Rosenbaum and Donald~B Rubin.
\newblock Assessing sensitivity to an unobserved binary covariate in an observational study with binary outcome.
\newblock \emph{Journal of the Royal Statistical Society: Series B (Methodological)}, 45\penalty0 (2):\penalty0 212--218, 1983.

\bibitem[Rubin(1974)]{rubin}
D.~B. Rubin.
\newblock Estimating causal effects of treatments in randomized and nonrandomized studies.
\newblock \emph{Journal of Educational Psychology}, 66\penalty0 (5):\penalty0 688, 1974.

\bibitem[Samothrakis et~al.(2022)Samothrakis, Matran-Fernandez, Abdullahi, Fairbank, and Fasli]{samothrakis22}
Spyridon Samothrakis, Ana Matran-Fernandez, Umar Abdullahi, Michael Fairbank, and Maria Fasli.
\newblock Grokking-like effects in counterfactual inference.
\newblock In \emph{2022 International Joint Conference on Neural Networks (IJCNN)}, pages 1--8. IEEE, 2022.

\bibitem[Taleb(2020)]{taleb20}
Nassim~Nicholas Taleb.
\newblock Statistical consequences of fat tails: Real world preasymptotics, epistemology, and applications.
\newblock \emph{arXiv preprint arXiv:2001.10488}, 2020.

\bibitem[Tan(2006)]{tan}
Zhiqiang Tan.
\newblock A distributional approach for causal inference using propensity scores.
\newblock \emph{Journal of the American Statistical Association}, 101\penalty0 (476):\penalty0 1619--1637, 2006.

\bibitem[Tibshirani et~al.(2019)Tibshirani, Foygel~Barber, Candes, and Ramdas]{tibshirani}
Ryan~J Tibshirani, Rina Foygel~Barber, Emmanuel Candes, and Aaditya Ramdas.
\newblock Conformal prediction under covariate shift.
\newblock \emph{Advances in neural information processing systems}, 32, 2019.

\bibitem[Wainwright(2019)]{wainwright19}
Martin~J Wainwright.
\newblock \emph{High-dimensional statistics: A non-asymptotic viewpoint}, volume~48.
\newblock Cambridge university press, 2019.

\bibitem[Wang et~al.(2022)Wang, Lyu, Wu, Wu, and Chen]{wang22}
Xin Wang, Shengfei Lyu, Xingyu Wu, Tianhao Wu, and Huanhuan Chen.
\newblock Generalization bounds for estimating causal effects of continuous treatments.
\newblock In Alice~H. Oh, Alekh Agarwal, Danielle Belgrave, and Kyunghyun Cho, editors, \emph{Advances in Neural Information Processing Systems}, 2022.

\bibitem[Wild et~al.(2023)Wild, Ghalebikesabi, Sejdinovic, and Knoblauch]{wild23}
Veit~David Wild, Sahra Ghalebikesabi, Dino Sejdinovic, and Jeremias Knoblauch.
\newblock A rigorous link between deep ensembles and (variational) bayesian methods.
\newblock \emph{arXiv preprint arXiv:2305.15027}, 2023.

\bibitem[Yao et~al.(2021)Yao, Chu, Li, Li, Gao, and Zhang]{yao}
Liuyi Yao, Zhixuan Chu, Sheng Li, Yaliang Li, Jing Gao, and Aidong Zhang.
\newblock A survey on causal inference.
\newblock \emph{ACM Transactions on Knowledge Discovery from Data (TKDD)}, 15\penalty0 (5):\penalty0 1--46, 2021.

\bibitem[Yin et~al.(2022)Yin, Shi, Wang, and Blei]{yin22}
Mingzhang Yin, Claudia Shi, Yixin Wang, and David~M Blei.
\newblock Conformal sensitivity analysis for individual treatment effects.
\newblock \emph{Journal of the American Statistical Association}, pages 1--14, 2022.

\end{thebibliography}
\newpage

\appendix

\section{Algorithm}\label{app:algo}

\begin{algorithm2e} 
  \SetKwInOut{Input}{Input}\SetKwInOut{Output}{Output}
  \caption{Greedy Quantile Maximizer \hfill (minimizer version is trivial) \label{alg:minimax}} 
  \Input{~Quantile rank $\beta$, weight bounds $(\underline\omega, \overline\omega)$ like those described in Assumption~\ref{ass:weigh}, and invertible cumulative density functions $F_1(y), F_2(y), \dots, F_n(y)$, which can be considered the conditional prediction distributions from the ensemble.}
  \Output{~Ensemble's $\beta$-quantile, $q\coloneqq \sup_w F^{-1}(\beta)$.}
  \vspace{0.25em}
  Initialize $w_i \gets 1$ for all $i=1,2,\dots n$\; 
  Compute initial search bounds $\underline q \gets \min_i F^{-1}_i(\beta)$ and $\overline q \gets \max_i F^{-1}_i(\beta)$\;
  \While{not converged}{
    Binary-search for $q\gets F^{-1}(\beta) \in(\underline q, \overline q)$, where $F(y) \coloneqq n^{-1}\sum_i w_i F_i(y)$\;
    Compute masses $\alpha_i \coloneqq F_i(q)$ for every $i$ and sort in ascending order (without relabeling)\;
    Find receiver~$r \coloneqq \arg\min_i \alpha_i$ such that $w_i<\overline\omega$\;
    Find sender~~~$s \coloneqq \arg\max_i \alpha_i$ such that $w_i>\underline\omega$\;
    \If{$r\geq s$}{
      \textbf{break}\;
    }
    Compute receivable $a \coloneqq \overline\omega - w_r$ and sendable $b \coloneqq w_s - \underline\omega$\;
    \eIf{$a < b$}{
      Transfer $w_r \gets \overline\omega$ and $w_s \gets w_s - a$\;
    }{
      Transfer $w_s \gets \underline\omega$ and $w_r \gets w_r + b$\;
    }
    Refine search bounds $\underline q \gets q$\;
  }
\end{algorithm2e}

This algorithm has quadratic asymptotic runtime in the ensemble size. As the bottleneck tends to be the quantile-search subroutine, one may benefit from implementing a \emph{bulk} weight-transfer procedure using Algorithm~\ref{alg:minimax} as a starting point.
\section{Proofs}\label{app:proofs}

\begin{proof}[Lemma~\ref{lem:finite-sample}]
  We study estimation errors in the weights in a manner inspired by Theorem~3 (supplementary) of \cite{lei21}. In our case, we study the cumulative distribution function (CDF) estimate\vspace{-1em}
  \begin{equation*}
    \hat F_{\tilde\omega}(y) = m^{-1}\sum_{j=1}^m \tilde \omega_j F_j(y) = m^{-1}\sum_{j=1}^m \tilde\omega(\Theta^{(j)}) F(y;\, \Theta^{(j)})
  \end{equation*}
  for $\hat p_\mathcal{M}(y_t\mid x; \mathcal{D}) \,=\, \hat\E_m[\,\tilde\omega(\Theta,t,x)\times p_\mathcal{M}(y\mid t,x;\Theta) \mid \mathcal{D}\,]$ with $t$ and $x$ omitted for brevity. We wish to accurately predict $Y_t\sim p_\mathcal{M}(Y_t\mid X=x;\mathcal{D})$. 
  Our main tool will be Hoeffding's inequality~\citep{wainwright19} in this endeavor. Namely, for any $u>0$,
  \begin{align*}
    \mathbb{P}[m\hat F_{\tilde\omega}(y) - \E m\hat F_{\tilde\omega}(y) \geq u] \leq \exp{-2u^2/m}
  \end{align*}
  because the individual CDFs are independent conditional on a fixed $y$, and take the range $[0,1]$. We focus on the upper bound (with quantile $1-\alpha/2$) of the prediction interval first, and extend that result to the lower bound by a symmetry argument.

  Resolving the expectation and factoring out $m$, we find
  \begin{equation}\label{eq:lemma-hoeffding}
    \mathbb{P}[\hat F_{\tilde\omega}(y) - F_{\tilde\omega}(y) \geq u] \leq \exp{-2mu^2} \quad \text{where } F_{\tilde\omega}(y) = \E[ \tilde\omega F(y;\,\Theta)].
  \end{equation}
  We observe that $F_{\tilde\omega}(y) = \E[ \omega F(y;\,\Theta)] + \E[ (\tilde\omega-\omega) F(y;\,\Theta)] =F_{\omega}(y) + F_{\tilde\omega-\omega}(y)$. Now let $u = 1 - F_{\tilde\omega}(y) + \alpha/2$ for the upper bound. This implies $\mathbb{P}[\hat F_{\tilde\omega}(y) \geq 1-\frac{\alpha}{2}] \leq \exp{-2m[1 - F_{\tilde\omega}(y) + \frac{\alpha}{2}]^2}$. Now we introduce the margin $\varepsilon>0$ and note that when $F_{\tilde\omega}(y) +\varepsilon\leq 1-\frac{\alpha}{2}$, then $u>0$ and we have
  \begin{equation*}
    \mathbb{P}\Big[\hat F_{\tilde\omega}(y) \geq 1-\frac{\alpha}{2}\Big] \leq \exp{-2m\varepsilon^2} \iff \mathbb{P}\Big[\hat F_{\tilde\omega}(y) < 1-\frac{\alpha}{2}\Big] > 1-\exp{-2m\varepsilon^2}.
  \end{equation*}
  Plugging in $y\coloneqq Y_t$ from the test set defined above, we find that this law is satisfied when $F_\omega(Y_t)\leq 1 - (\frac{\alpha}{2} + \varepsilon + F_{\tilde\omega-\omega}(Y_t))$. The fact that $F_\omega(Y_t)$ is uniformly distributed (following Assumption~\ref{ass:weigh}) allows us to conclude that the condition is met with probability at least $1-(\frac{\alpha}{2} + \varepsilon + \E\abs{\tilde\omega-\omega})$, observing the triangle inequality of the absolute norm.

  Applying the same reasoning to the prediction interval's lower bound eventually yields
  \begin{equation*}
    \mathbb{P}\Big[\frac{\alpha}{2} < \hat F_{\tilde\omega}(Y_t) < 1-\frac{\alpha}{2}\Big] > 1-2\exp{-2m\varepsilon^2}.
  \end{equation*}
  with aggregate probability $1-\beta'$ for $\beta'=\alpha + 2\varepsilon + 2\E\abs{\tilde\omega-\omega}$. Applying the inverse CDF and substituting $\varepsilon\coloneqq\varepsilon/2$ (without loss of generality), we obtain the final form $\beta=\alpha + \varepsilon + 2\E\abs{\tilde\omega-\omega}$; hence with probability $1-\beta$,
  \begin{equation*}
    \mathbb{P}\Big[Y_t \in \Big(\hat F_{\tilde\omega}^{-1}\big(\alpha/2\big),\, \hat F_{\tilde\omega}^{-1}\big(1-\alpha/2\big)\Big)\Big] > 1-2\exp{-m\varepsilon^2/2}.
  \end{equation*}
\end{proof}

\begin{proof}[Theorem~\ref{thm:valid}]
  Here we extend Lemma~\ref{lem:finite-sample} to partially identifiable weights that yield an admissible set of prediction intervals. The outcome interval under consideration is the union of all these admissible intervals. This is attained by a supremum over the upper bound and an infimum over the lower bound. 
  In applying Hoeffding's bound on the upper bound, we can replace Equation~\ref{eq:lemma-hoeffding} with
  \begin{equation*}
    \mathbb{P}[\hat F_{\tilde\omega^+}(y) - F_{\tilde\omega^+}(y) \geq u] \leq \exp{-2mu^2} \quad \text{such that } \hat F_{\tilde\omega^+}(y) = \sup \hat F_{\tilde\omega}(y).
  \end{equation*}
  Additionally, we define $\omega^+$ to satisfy $\hat F_{\omega^+}(y) = \sup \hat F_{\omega}(y)$. Let $\beta=\alpha + \varepsilon + \E\abs{\tilde\omega^+-\omega^+} + \E\abs{\tilde\omega^- -\omega^-}$ where $(\tilde\omega^-,\omega^-)$ are analogously defined for their respective infima. By a straightforward extension of Lemma~\ref{lem:finite-sample}, we have Equation~\ref{eq:coverage} with probability at least $1-\beta$. By the assumption stated in this theorem, we know that $\tilde\omega^{\pm}(\Theta) \in \{\underline{\tilde\omega}, \overline{\tilde\omega}\}$ almost surely over the weight assignments. It is clear, then, that $\E\abs{\tilde\omega^+-\omega^+} + \E\abs{\tilde\omega^- -\omega^-} \leq 2\E[\,\abs{\underline{\tilde\omega} - \underline{\omega}} \vee \abs{\overline{\tilde\omega} - \overline{\omega}}\,]$, completing the proof.
\end{proof}

\begin{proof}[Theorem~\ref{thm:opt}]
  The putative optimality condition for the maximization problem solved by Algorithm~\ref{alg:minimax}, restated, is for there to be no pair of mixture components $(\theta^{(j)}, \theta^{(k)})$ such that $\omega(\theta^{(j)}) > \underline\omega$ and $\omega(\theta^{(k)}) < \overline\omega$, as well as $F(q;\theta^{(j)}) > F(q;\theta^{(k)})$ where $q$ is the current $\beta$-quantile: 
  \begin{equation*}
    \beta = F(q) \triangleq m^{-1}\sum_i \omega(\theta^{(i)}) F(q; \theta^{(i)}).
  \end{equation*}
  
  We will prove both directions of entailment to establish equivalence. First, we must show that if the quantile is maximized, then the condition holds. Suppose that $q$ is the maximal quantile under the problem constraints and the condition is not satisfied, so there indeed is a pair $(\theta^{(j)}, \theta^{(k)})$ as described. This implies that there is weight that could be transferred, of the amount
  \begin{equation*}
    \Delta \omega \triangleq \min\{ \omega(\theta^{(j)}) - \underline\omega,\ \overline\omega - \omega(\theta^{(k)}) \} > 0.
  \end{equation*}
  Transferring that weight would yield a new mixture
  \begin{equation*} 
    G(\cdot)=\frac{1}{m}\sum_i\omega(\theta^{(i)})F(\cdot\,; \theta^{(i)}) + \frac{\Delta\omega}{m}[F(\cdot\,; \theta^{(k)}) - F(\cdot\,; \theta^{(j)})],
  \end{equation*}
  with the consequence of $G(q)<F(q)$ because $F(q;\theta^{(j)}) > F(q;\theta^{(k)})$. Therefore $G^{-1}(\beta)> q$ due to monotonicity and $q$ is not the optimal quantile. By contraposition, optimality entails our stated optimality condition. Now for the converse.

  With similar notation as above, we have $F(q)=\beta$ but come into the posession of some feasible $G(\cdot)$ where $G(q^*)=\beta$ and $q^*>q,$ so $q$ is no longer optimal. Deconstruct $G(\cdot) = m^{-1}\sum_i\omega'(\theta^{(i)})F(\cdot\,; \theta^{(i)})$. By monotonicity, $G(q)<F(q)$. Hence
  \begin{equation*}
    m^{-1}\sum_i[\omega'(\theta^{(i)}) - \omega(\theta^{(i)})]F(\cdot\,; \theta^{(i)})<0.
  \end{equation*}

  Ignoring the identical pairs of weights between $F$ and $G$,
  \begin{multline*}
    \sum_{i \in \mathcal{A}} [\omega(\theta^{(i)}) - \omega'(\theta^{(i)})] F(q; \theta^{(i)}) > \sum_{i \in \mathcal{B}} [\omega'(\theta^{(i)}) - \omega(\theta^{(i)})] F(q; \theta^{(i)}),\\
    \mathcal{A}\triangleq\{i: \omega(\theta^{(i)}) > \omega'(\theta^{(i)})\},\quad \mathcal{B}\triangleq\{i: \omega'(\theta^{(i)}) > \omega(\theta^{(i)})\}.
  \end{multline*}
  At the same time, $\sum_{i \in \mathcal{A}} [\omega(\theta^{(i)}) - \omega'(\theta^{(i)})] = \sum_{i \in \mathcal{B}} [\omega'(\theta^{(i)}) - \omega(\theta^{(i)})]$ because of the constraint on the probability simplex.
  For the above inequality to be valid alongside this equality, there \emph{must} be at least one pair $(j\in\mathcal{A},\ k\in\mathcal{B})$ such that $F(q;\theta^{(j)}) > F(q;\theta^{(k)})$. Hence the negation of the optimality condition holds, and by contraposition we prove the other entailment direction.
\end{proof}

\newpage
\section{Experimental Details on \texttt{PBMC}}\label{app:pbmc}

The original PBMC dataset~\citep{kang} had 14,039 cells that were used as data points for our benchmark. In each of the 16 trials, we randomly allocated 8,192 ($2^{13}$) training instances, 2,048 ($2^{11}$) validation instances, and 2,048 ($2^{11}$) test insances from the original sample. Then we designed the causal system by projecting the cells' 17,796 gene expressions into vectors
\begin{equation*}
  V\triangleq\langle \text{32 visible confounders\dots, 1 treatment, 32 hidden confounders\dots} \rangle\in\mathbb{R}^{65}
\end{equation*}
by drawing 65 i.i.d normal coefficients. The 64 confounding entries were rank-normalized to give them $\text{Uniform}[0,1)$ marginals. The treatment entry was binarized by thresholding at $2/3$ so that the data were slightly unbalanced with more $(T=0)$ observations.
The outcome link was determined by a random matrix $M_{ij} \sim \text{i.i.d Normal}(0,1)$. The diagonal coefficient $M_{33,33}$ corresponding to the treatment entry was upscaled by a factor of 64 to keep the treatment effect discernible from the rest. The pre-noised outcome $U\triangleq V^\text{T} M V$ endured strong quadratic treatment and confounding effects. Finally, the observed outcome was $\text{Cauchy}(\mu=U, \sigma=1)$-distributed as motivated in \S\ref{sec:bench-pbmc}. We aimed to keep this mathematical construction of the semi-synthetic benchmark \emph{parsimonious} by introducing a minimal number of design choices and nontrivial default values.

\section{Experimental Details on \texttt{AITA}}\label{app:aita}

\begin{figure}[!ht]\centering
  \includegraphics[width=0.85\linewidth]{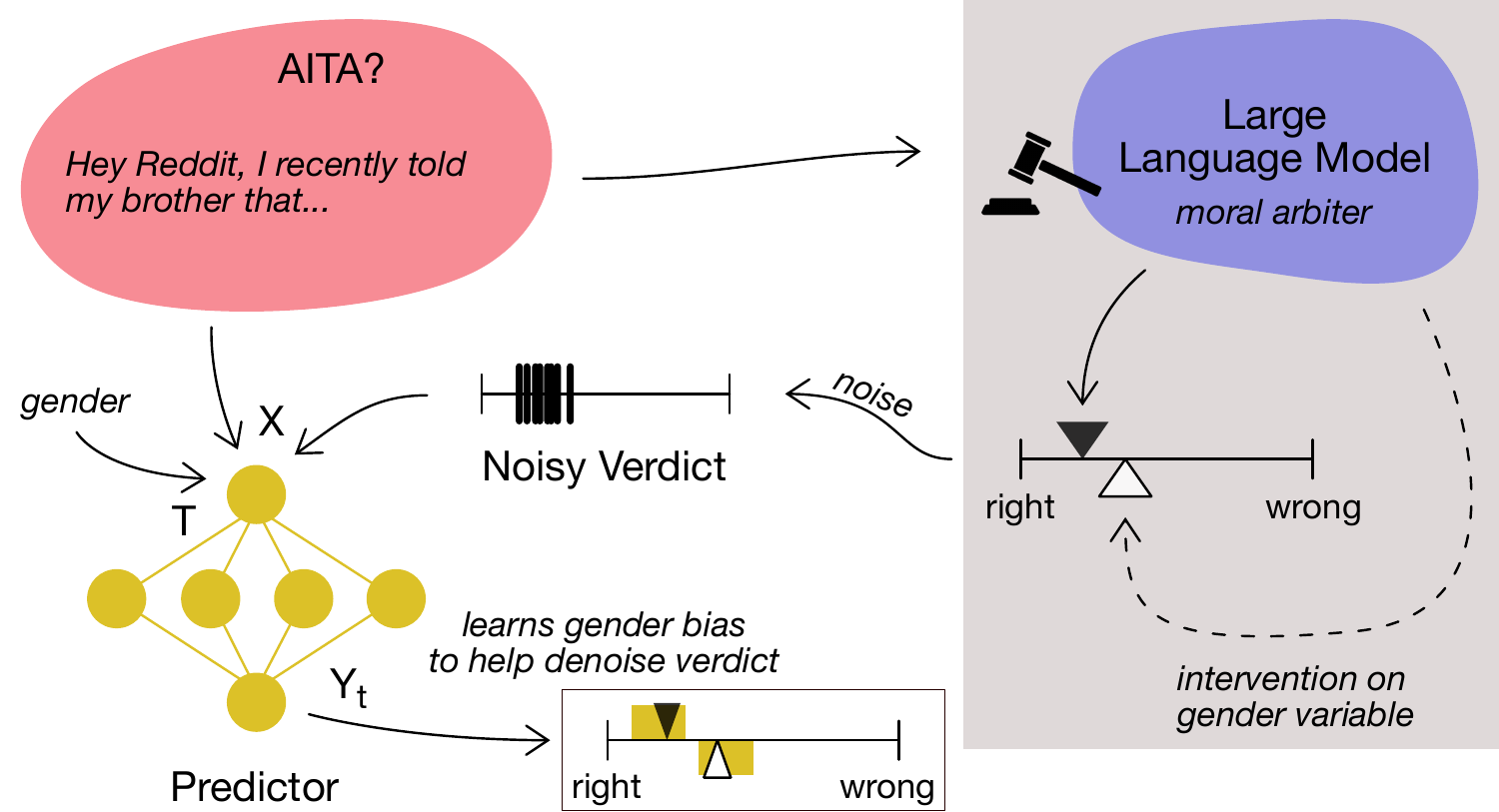}
  \vspace{0.25em}
  \caption{\label{fig:aita}Illustration for the \texttt{AITA} benchmark as a denoising task. The predictor receives a rough verdict through a noisy channel, relieving it from having to learn the moral judgment of GPT-4 and allowing it to focus on the biases due to gender and topics that help with denoising.}
\end{figure}

Figure~\ref{fig:aita} depicts the experimental setup. We used OpenAI's \texttt{gpt-4-0314} API model. The \citet{obrien20} dataset contained 11,862 posts with detectable gender identifiers, of which 9,000 were reserved for the observational data. 8,000 of those were for training and the other 1,000 for validation. The 2,862 posts that were not included in the observational dataset were used as the causal outcome test set by intervening on the gender variable before feeding them into GPT-4. Please refer to \texttt{scripts/aita-preprocessing.py} in the supplementary source for all the querying and preprocessing details. Processed data are included as CSVs. View the chat-prompt prologue below.

\begin{tabular}{l |@{\hspace{2em}} l}
  {System} & \parbox{.75\textwidth}{You are an average person acting as a subjective moral arbiter on r/AmITheAsshole.\\
  Whenever someone presents a story, you respond with a number from 1 to 99 where 1 is noble, 49 is completely neutral (rare), and 99 is atrocious.}
  \\\midrule
  {User} & I was mean for no reason.\\\midrule
  {Assistant} & 90\\
  \midrule
  {User} & I tipped the waiter more than usual.\\\midrule
  {Assistant} & 10\\
\end{tabular}

Unfortunately, we hit OpenAI's GPT-4 quota before we could evaluate all arms of the potential outcomes. It did not hinder the benchmark, but it did prevent thorough analysis of the causal system.

\section{Additional Results for \texttt{IHDP}}


\begin{table}[ht]\centering 
  \begin{tabular}{l | r r r } 
    Conformal Baseline & 50 - 50 Split & 75 - 25 Split & 87.5 - 12.5 Split* \\
    \midrule
    Ens-CSA-DCP & 1 Failure / 10 Trials & 2 Failures / 10 Trials & 5 Failures / 10 Trials \\
    Ens-CSA-CQR & 2 Failures / 10 Trials & 2 Failures / 10 Trials & 6 Failures / 10 Trials \\
  \end{tabular}\vspace{0.50em}

  \caption{\label{tab:ihdp-splits}Failure rates of the baseline methods applied to the \texttt{IHDP} benchmark with 95\% target coverage. The results in Figure~\ref{fig:ihdp-scatter} use the 50-50 split that appears to work best. Asterisk marks the arrangement where the entire original training set is used for estimation, and the validation set for calibration. 
  The other benchmarks (\texttt{PBMC}, \texttt{AITA}) have larger samples that obviate this issue. }
\end{table}

\section{Hardware Details}

All results were obtained on an Intel Xeon server using a single NVIDIA GeForce RTX 2080 Ti.

\section{Code Availability}
Please refer to our Julia package at\\\url{https://github.com/marmarelis/TreatmentCurves.jl}.

\end{document}